%% file: main.tex
\documentclass[11pt]{article}
\usepackage[margin=1in]{geometry}
\input{tex/macros}              

\title{Clip Your Sequences Fairly: \\Enforcing Length Fairness for Sequence‑Level RL} 
\author{
Hanyi Mao\textsuperscript{1}\thanks{ \texttt{hanyim@uchicago.edu}}
\quad Quanjia Xiao\textsuperscript{2}
\quad Lei Pang\textsuperscript{2}
\quad Haixiao Liu\textsuperscript{3}\\
\textsuperscript{1} University of Chicago\quad
\textsuperscript{2} Peking University\quad
\textsuperscript{3} Duxiaoman
}

\date{}

\begin{document}
\maketitle
\begin{abstract}

We propose \textbf{FSPO} (Fair Sequence Policy Optimization), a sequence-level reinforcement learning method for LLMs that enforces length-fair clipping on the importance-sampling (IS) weight. We study RL methods with sequence-level IS and identify a mismatch when PPO/GRPO-style clipping is transplanted to sequences: a fixed clip range systematically reweights short vs.\ long responses, distorting the optimization direction. FSPO introduces a simple remedy: we clip the sequence log-IS ratio with a band that scales as $\sqrt{L}$. Theoretically, we formalize length fairness via a Length Reweighting Error (LRE) and prove that small LRE yields a cosine directional guarantee between the clipped and true updates. Empirically, FSPO flattens clip rates across length bins, stabilizes training, and outperforms baselines across model sizes and evaluation datasets, with the largest gains on the Qwen3‑8B‑Base model.
\end{abstract}

\input{sections/intro}
\input{sections/lre}
\input{sections/gaussian}
\input{sections/method}
\input{sections/experiments}

\input{sections/results}
\input{sections/conclusion}

\bibliographystyle{plainnat}
\newpage
\bibliography{bibs/references}  

\newpage
\appendix
\input{sections/appendix}
\end{document}

%% file: tex/macros.tex
\usepackage[T1]{fontenc}
\usepackage[utf8]{inputenc}
\usepackage{lmodern}
\usepackage{microtype}

\usepackage{amsmath,amssymb,amsthm,mathtools}
\usepackage{bm}
\usepackage{bbm}       
\usepackage{dsfont}    

\newtheorem{theorem}{Theorem}[section]
\newtheorem{assumption}{Assumption}[section]
\newtheorem{definition}{Definition}[section]

\DeclareMathOperator{\KL}{KL}

\newcommand{\EE}{\mathbb{E}}

\newcommand{\clip}{\mathrm{clip}}

\usepackage{graphicx}
\usepackage{subcaption}
\usepackage{booktabs}
\usepackage{caption}
\graphicspath{{figs/}}
\usepackage{float}
\usepackage[section]{placeins}

\usepackage[numbers,sort&compress]{natbib}
\usepackage{xcolor}
\definecolor{linkcol}{RGB}{20,70,140}
\usepackage[
  colorlinks=true,
  linkcolor=linkcol,
  citecolor=linkcol,
  urlcolor=linkcol
]{hyperref}
\usepackage[nameinlink,capitalize,noabbrev]{cleveref}

\crefname{theorem}{Theorem}{Theorems}
\Crefname{theorem}{Theorem}{Theorems}
\crefname{assumption}{Assumption}{Assumptions}
\Crefname{assumption}{Assumption}{Assumptions}
\crefname{definition}{Definition}{Definitions}
\Crefname{definition}{Definition}{Definitions}
\crefname{lemma}{Lemma}{Lemmas}
\Crefname{lemma}{Lemma}{Lemmas}
\crefname{proposition}{Proposition}{Propositions}
\Crefname{proposition}{Proposition}{Propositions}
\crefname{remark}{Remark}{Remarks}
\Crefname{remark}{Remark}{Remarks}

\newcommand{\cX}{\mathcal{X}}

%% file: sections/intro.tex
\section{Introduction}
\label{sec:intro}

Recent progress on reinforcement learning (RL) for large language models (LLMs) has been catalyzed by GRPO~\citep{grpo} and the broader RLVR paradigm~\citep{lambert2025tulu3}, where rule-based, verifiable rewards are assigned to the entire response rather than token-wise signals. This framing has proven effective for improving mathematical reasoning and other verifiable tasks \citep{deepseekai2025deepseekr1incentivizingreasoningcapability, rlvr-more, wang2025oneshotrlvr}. However, the optimization procedures used in current RLVR systems largely inherit token-level machinery from PPO-like methods~\citep{schulman2017ppo}, including the use of token-level importance-sampling (IS) ratios and token-level clipping. Meanwhile, subsequent works emphasize that once rewards are sequence-level, it is more faithful to operate with sequence-level IS so as to match the reward granularity~\citep{rloo,gspo}.

Despite the shift toward sequence-level IS, the theoretical distinctions and practical consequences of clipping in this setting remain underexplored. Existing sequence-level IS methods~\citep{rloo,gspo} transplant the clipping mechanism from token-level methods directly and apply a \emph{fixed} clip range to the probability ratio of the whole sequence. We argue that fixed sequence-level clipping is problematic: the dispersion of sequence-level \emph{log} ratios increases with response length \(L\). A fixed band therefore induces length-dependent acceptance rates and systematically reweights short versus long responses.

This paper studies sequence-level clipping through the lens of \emph{length fairness}. We formalize a simple criterion: \emph{acceptance rates should be approximately constant across response lengths}. We show that fixed sequence-level clipping violates this criterion and can distort the training target. To address this, we propose \textbf{FSPO} (Fair Sequence Policy Optimization). FSPO preserves IS semantics and restores length fairness by using a \(\sqrt{L}\)-scaled acceptance band on the sequence log-ratio, which approximately equalizes acceptance across lengths.

To ground our analysis, we evaluate \textbf{FSPO} on sequence-level RL for mathematical reasoning. We compare against two sequence-level baselines: (i) \emph{RLOO} with sequence-level IS and a fixed clip on the full-sequence ratio, and (ii) \emph{GSPO} with ratio normalization. We report Avg@8 on \textsc{MATH500}, Avg@32 on \textsc{AIME24} and \textsc{AIME25}, alongside diagnostic plots that measure acceptance rate as a function of response length to verify length fairness. Across two base model scales, we observe flatter acceptance across length bins, more stable training dynamics, and improved task scores; full results and ablations are presented in \Cref{sec:results}.

Background on RL for LLMs and RLVR is provided in Appendix \ref{app:prelim}, and a detailed justification for sequence-level IS weights in RLVR scenario is given in Appendix \ref{app:why_seq}.

%% file: sections/lre.tex
\section{Length Fairness and Length Reweighting Error (LRE)}
\label{sec:lre}

Following the notation in Appendix \ref{app:prelim}, define the sequence-level log importance ratio
\[
S(y \mid x)\;=\;\log \pi_{\theta}(y \mid x)\;-\;\log \pi_{\theta_{\text{old}}}(y \mid x).
\]
Let \(L=\mathrm{len}(y)\) and fix a length-indexed band \(b_L>0\). The acceptance event (unclipped) is defined as
\[
\mathcal A_L \;=\; \big\{\,y \,\bigm|\, \lvert S(y\mid x)\rvert \le b_L \,\big\}.
\]
We define the length-conditional acceptance rate \(q(L)=\Pr_{\pi_{\theta_{\text{old}}}}(\mathcal A_L \mid L)\), and the per-length contributions
\begin{align*}
    g_L^\star \;&=\; \EE_{\pi_{\theta_{\text{old}}}}\!\Big[\frac{\pi_{\theta}(y\mid x)}{\pi_{\theta_{\text{old}}}(y\mid x)}\,\nabla_{\theta}\log \pi_{\theta}(y\mid x)\,A(x,y)\,\Bigm|\,L\Big],\quad\\
g_L^{\,b} \;&=\; \EE_{\pi_{\theta_{\text{old}}}}\!\Big[\frac{\pi_{\theta}(y\mid x)}{\pi_{\theta_{\text{old}}}(y\mid x)}\,\nabla_{\theta}\log \pi_{\theta}(y\mid x)\,A(x,y)\,\Bigm|\,\mathcal A_L,\,L\Big],
\end{align*}

so that the true policy gradient target and its clipped surrogate are
\[
g^\star\;=\;\EE_{L}\big[g_L^\star\big],\qquad
g^{\,b}\;=\;\EE_{L}\big[q(L)\,g_L^{\,b}\big].
\]

\begin{definition}[Length Reweighting Error (LRE)]
Let \(\bar q=\EE\!\big[q(L)\big]\). Define
\[
\mathrm{LRE}\;=\;\tfrac12\,\EE\!\left[\left|\frac{q(L)}{\bar q}-1\right|\right].
\]
\end{definition}

Small LRE means the acceptance rate is nearly constant across response lengths.  

Let \(\kappa=\dfrac{\EE\!\big[\lVert g_L^\star\rVert\big]}{\lVert g^\star\rVert}\ge 1\), which captures the dispersion of per-length signal magnitude.

\begin{assumption}[Bounded stratification]\label{ass:weak-strat}
There exists \(\eta\in[0,1)\) such that for all \(L\),
\[
\big\lVert g_L^{\,b}-g_L^\star \big\rVert \;\le\; \eta\,\lVert g_L^\star\rVert .
\]
\end{assumption}
This assumption states that clipping does not severely distort the target within each length stratum.

\begin{assumption}[Bounded correlation]\label{ass:cov}
The correlation between \(\lvert q(L)-\bar q\rvert\) and \(\lVert g_L^\star\rVert\) is mild so that
\[
\EE\!\Big[\,\lvert q(L)-\bar q\rvert\,\lVert g_L^\star\rVert\,\Big]
\;\le\;
\gamma\,\EE\!\big[\lvert q(L)-\bar q\rvert\big]\;\EE\!\big[\lVert g_L^\star\rVert\big].
\]
\end{assumption}
This assumption is optional; see Appendix \ref{app:angle-proof}.

\begin{theorem}[Directional guarantee under length fairness]\label{thm:angle}
Under \Cref{ass:weak-strat,ass:cov},
\[
\cos\angle\!\big(g^{\,b},\,g^\star\big)\ \ge\ \frac{1-\rho}{1+\rho},\qquad
\rho\ \le\ \kappa\Big(\eta+2\gamma(1+\eta)\,\mathrm{LRE}\Big).
\]
\end{theorem}

The theorem implies that smaller LRE yields a larger lower bound on the cosine similarity between the clipped update and the true update. The proof and further discussion are provided in Appendix \ref{app:angle-proof}.

%% file: sections/gaussian.tex
\section{Distribution of Sequence-Level Log Ratio}
\label{sec:gaussian}

In this section we study the distribution of the sequence-level log importance-sampling (IS) ratio and derive practical guidance for designing procedures that achieve the \emph{length fairness} criterion introduced earlier.

We view decoding under an LLM \(\pi\) with a limited context window \(K\) as a finite-state Markov chain on \(V^K\), where \(V\) is the vocabulary; this reduction for autoregressive language models is discussed by \citet{llm_markov} in detail. Under randomized sampling with nonzero temperature, the chain is irreducible and aperiodic. Therefore, the Markov CLT for additive functionals~\citep{jones2004mclt,maxwell2000additiveclt} applies to
\[
S_L \;=\; \sum_{t\le L}\log\frac{\pi_{\theta}(y_t\mid h_t)}{\pi_{\theta_{\text{old}}}(y_t\mid h_t)}\,,
\]
which yields the following theorem:

\begin{theorem}[Gaussianity of the sequence-level log ratio]\label{thm:clt}
The sequence log-IS ratio obeys an asymptotically Gaussian law:
\[
\frac{S_L-\mu_L}{\sqrt{L}}\ \Rightarrow\ \mathcal{N}(0,\sigma^2),
\qquad \mu_L=\\theta(L),\quad \sigma^2>0.
\]
\end{theorem}

\begin{figure}[t]
  \centering
  \includegraphics[width=0.48\linewidth]{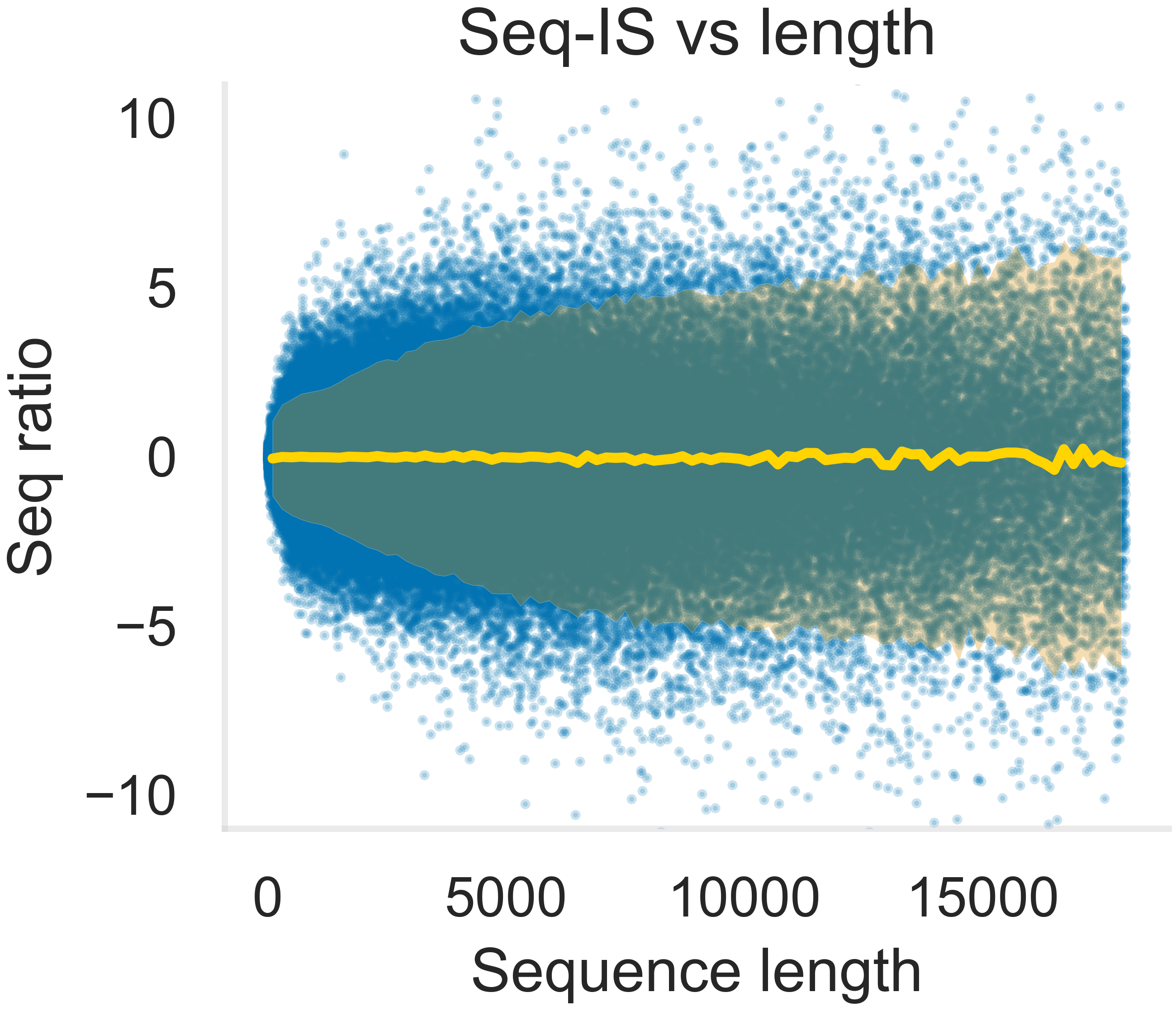}\hfill
  \includegraphics[width=0.48\linewidth]{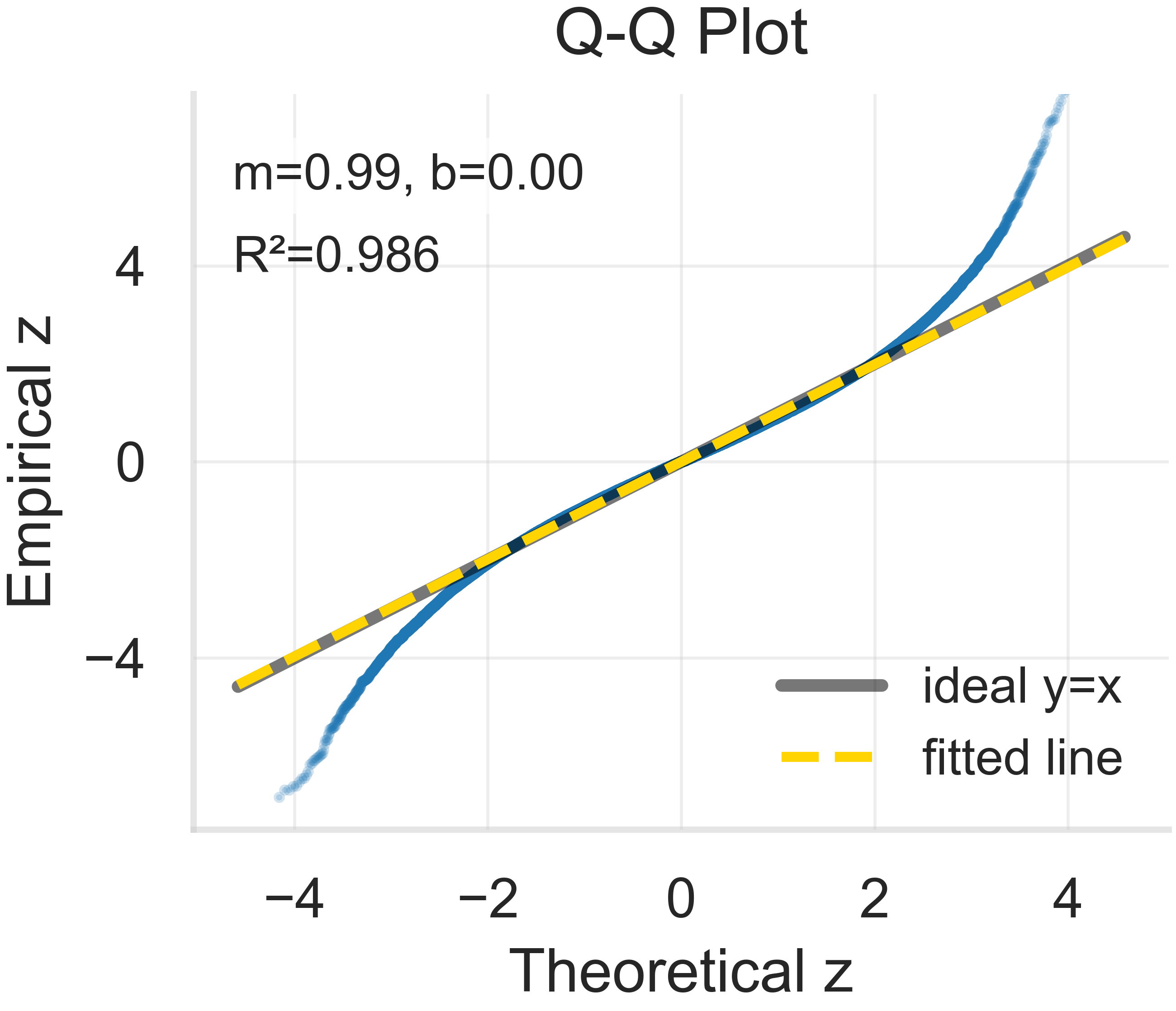}
  \caption{\textbf{Empirical analysis of the sequence-level IS ratio.} Sample size \(n=217{,}454\).
  \textbf{Left:} Empirical distribution of \(S_L\). The yellow line shows the empirical mean and the shaded band the \(\pm2\) empirical standard deviation, computed with a bin size of 200 (see Appendix \ref{app:angle-proof} for justification of binning).
  \textbf{Right:} Q–Q plot testing normality. The sorted data point quantiles are shown in blue dots. We report the slope \(m\), intercept \(b\), and \(R^2\) of the fitted line.}
  \label{fig:seq_ratio_qqplot}
\end{figure}

\Cref{fig:seq_ratio_qqplot} illustrates the empirical distribution of the sequence-level log IS ratio using all steps across the full training run. Consistent with the theorem, the empirical standard deviation of \(S_L\) grows approximately \(\propto \sqrt{L}\). The observed estimator is \(\hat\sigma = 0.0304\).

To further assess normality, we compute the standardized statistic
\[
\hat Z \;=\; \frac{S_L-\hat\mu_L}{\sqrt{L}\,\hat\sigma},
\]
where \(\hat\mu_L\) is computed within each length bin and \(\hat\sigma\) is estimated from all values of \((S_L-\hat\mu_L)/\sqrt{L}\). The Q–Q plot (right panel) shows that the fitted line coincides with the \(y=x\) reference; the empirical distribution exhibits slightly heavier tails, but within \(\pm2\) standard deviations it is very close to normal.

In \Cref{fig:seq_ratio_qqplot} (left), the estimated per-length mean \(\hat\mu_L\) exhibits slightly larger fluctuations at larger lengths but remains small relative to \(\hat\sigma\), thus empirically we set \(\hat\mu_L\approx 0\).

\textbf{Theoretical clip-fraction patterns of RLOO and GSPO.}
By \Cref{thm:clt}, \(S_L \approx \mathcal N(\mu_L,\sigma^2 L)\).
Let \(\Phi(\cdot)\) be the standard normal CDF.  Similar to the acceptance-rate notation \(q(L)\) used in \Cref{sec:lre}, We denote the \emph{clip probability} by \(c(L) \coloneq 1 - q(L)\). For a symmetric two-sided clip in log space:
\begin{align}
\textbf{RLOO:}\quad
c_{\mathrm{RLOO}}(L)
&= \Pr\!\big(|S_L|>\xi\big)
= 2\,\Phi\!\left(-\,\frac{\xi-\mu_L}{\sigma\sqrt{L}}\right)
\;\approx\; 2\,\Phi\!\left(-\,\frac{\xi}{\sigma\sqrt{L}}\right), \label{eq:c_rloo}
\\[2mm]
\textbf{GSPO:}\quad
c_{\mathrm{GSPO}}(L)
&= \Pr\!\big(|S_L|>\xi L\big)
= 2\,\Phi\!\left(-\,\frac{\xi L-\mu_L}{\sigma\sqrt{L}}\right)
\;\approx\; 2\,\Phi\!\left(-\,\frac{\xi\sqrt{L}}{\sigma}\right), \label{eq:c_gspo}
\end{align}
where the approximations use the empirically small drift \(\mu_L\approx 0\).
Both schemes induce clip probabilities that vary systematically with \(L\).

To obtain a \emph{constant} (length-independent) clip probability, \textbf{FSPO} sets
\(b_L=\mu_L+z\,\sigma\,\sqrt{L}\).
With the same calculation as in \Cref{eq:c_rloo,eq:c_gspo}, we obtain
\[
c_{\mathrm{FSPO}}(L)\;\approx\; 2\,\Phi(-z),
\]
which is independent of \(L\) and hence preserves the length fairness required by \Cref{thm:angle}. Moreover, as suggested by the Q–Q plot in \Cref{fig:seq_ratio_qqplot} (right), choosing \(z<2\) keeps us in a regime where the normal approximation is highly accurate.

We plot the theoretical clip–probability curves in \Cref{fig:qL}, together with the empirically observed clip fractions, showing close agreement with the theory.

%% file: sections/method.tex
\section{Method: FSPO}
\label{sec:method}

For each prompt \(x \sim \mathcal{D}\) we sample \(G\) completions
\(\{y_i\}_{i=1}^G \sim \pi_{\theta_{\text{old}}}(\cdot\mid x)\) and optimize the PPO-style pessimistic surrogate
\begin{equation}
\label{eq:obj-fspo}
\mathcal{J}_{\text{FSPO}}(\theta)
\;=\;
\EE_{x,\,\{y_i\}}\!\left[
\frac{1}{G}\sum_{i=1}^{G}
\min\!\Big\{\exp\!\big(S_{\theta}(y_i\mid x)\big)\,\widehat{A}_i\ ,\
\exp\!\big(\clip\big(S_{\theta}(y_i\mid x),-b_{L_i},\,b_{L_i}\big)\big)\,\widehat{A}_i
\Big\}
\right],
\end{equation}
where \(\widehat{A}_i\) is an advantage estimate and \(\clip(s,\ell,u)=\min\{\max\{s,\ell\},u\}\).
The sequence-level log importance ratio is
\begin{equation}
\label{eq:Si}
S_{\theta}(y_i\mid x)
\;=\;
\log \frac{\pi_{\theta}(y_i\mid x)}{\pi_{\theta_{\text{old}}}(y_i\mid x)}
\;=\;
\sum_{t=1}^{L_i}\log\frac{\pi_{\theta}(y_{i,t}\mid h_{i,t})}{\pi_{\theta_{\text{old}}}(y_{i,t}\mid h_{i,t})},
\end{equation}
with \(L_i=\mathrm{len}(y_i)\) and \(h_{i,t}=(x,y_{i,<t})\) the prefix at step \(t\).
FSPO performs \emph{log-space} clipping by truncating \(S_{\theta}\) to \([{-}b_L,b_L]\) before exponentiation, using the band as discussed in \Cref{sec:gaussian}
\begin{equation}
\label{eq:band}
b_{L}
\;=\;
\underbrace{\hat{\mu}_L}_{\text{drift}}
\;+\;
\underbrace{z\,\hat{\sigma}\,\sqrt{L}}_{\text{scale}}.
\end{equation}
Note that in \Cref{eq:obj-fspo} we average over the number of sequences $G$. This is natural for token-level clipping, but at sequence-level clipping is applied to the entire sequence and the clip fraction is typically much larger. A natural idea would be to exclude clipped sequences from the average. However, we keep $G$ as the denominator, which serves as a dynamic step-size adjustment: when the clip fraction is higher which indicates that current mini-batch is unstable with higher variance, keeping the denominator at $G$ correspondingly yields a smaller effectivem update for that batch.

\textbf{Drift term.}
Following \Cref{sec:gaussian}, we set \(\hat\mu=0\).
We empirically validate this simple setting flattens the clipping fraction and improves performance. In fact, the drift connects to token-level KLs:
\begin{equation}
\label{eq:mu_kl}
\EE_{\pi_{\theta_{\text{old}}}}[S_L]
\;=\;
\EE_{\pi_{\theta_{\text{old}}}}\!\Bigg[\sum_{t=1}^{L} \log\frac{\pi_{\theta}(y_t\mid h_t)}{\pi_{\theta_{\text{old}}}(y_t\mid h_t)}\Bigg]
\;=\;
\sum_{t=1}^{L} -\KL\!\big(\pi_{\theta_{\text{old}}}(\cdot\mid h_t)\,\|\,\pi_{\theta}(\cdot\mid h_t)\big).
\end{equation}
This further justifies our setting of $\hat\mu=0$, as we observe the KL between old and new policies is very small. However, this might not hold true for experimental settings significantly different than ours, in which cases a refined drift treatment could leverage \eqref{eq:mu_kl}; we leave this for future work.

\textbf{Scale term.}
In \eqref{eq:band}, \(z\) controls the target clip fraction (\Cref{sec:gaussian}), while \(\hat\sigma\) is tracked by a running estimator over recent batches.
In practice we set \(c \coloneqq z\,\hat\sigma\) and tune \(c\) as a single hyperparameter.
A natural extension is to use asymmetric scales $c_{\mathrm{upper}}$ and $c_{\mathrm{lower}}$, allowing separate control of the upper and lower clip ranges as in \citet{dapo}.
Current RL implementations commonly include \emph{dual-clip} \citep{dualclip}, which effectively clips the ratio at $(1+\epsilon_{\text{dual}})$ when $A<0$; in our experiments, we also implement dual-clip in log-space and tune $c_{\text{dual}}$.
Implementation details and discussion on hyperparameter tuning are provided in \Cref{app:hyperparam_tuning}.

\textbf{Compatibility with other components}
FSPO is a lightweight, plug-in modification that only changes the importance-ratio term in the policy loss.
To isolate its effect, our implementation keeps all remaining components identical to the baselines (e.g., GRPO-style advantage).
FSPO is compatible with alternative advantage estimators (e.g., \(\widehat A^{\mathrm{LOO}}\)~\citep{rloo19,drgrpo}), data filtering~\citep{dapo}, and overlength penalties~\citep{dapo}, among others.

%% file: sections/experiments.tex
\section{Experimental Setup}
\label{sec:experiments}

\subsection{Models and data.}
We evaluate our method on two base LLMs: \textbf{Qwen3-1.7B-Base} and \textbf{Qwen3-8B-Base} \cite{qwen3}.
For training, we use DAPO-Math-17K~\cite{dapo} together with AIME problems up to and including 2023 \cite{maa_aime_1983_2023}, accessed from \cite{aime_1983_2024}.
Evaluation is conducted on held-out math benchmarks: MATH500 \cite{math500}, AIME24 \cite{aime24}, and AIME25 \cite{aime25}. We did not include MATH500 training set as we argue that its difficulty is inadequate for efficient training.
We report \textbf{Avg@8} (per-sample accuracy averaged over 8 sampled completions) on MATH500, and \textbf{Avg@32} on AIME24/AIME25, since each AIME set contains only 30 questions and \textbf{Avg@32} yields more stable estimates. The detailed sampling configurations of evaluation process are provided in Appendix \ref{app:config}.

\subsection{Training framework.}
We build on \textsc{VeRL}~\cite{verl} with vLLM \cite{vllm} as roll-out backend and Megatron-LM \cite{megatron-lm} as training backend.
All models are trained under identical sampling policies, batch sizes, and total token budgets.
Complete hyperparameters and infrastructure details are provided in Appendix~\ref{app:config}.

\subsection{Baselines.}
We compare against sequence-level RL baselines: \textbf{RLOO}, \textbf{GSPO}, and our \textbf{FSPO}.
We also include \textbf{GRPO} as an important baseline to highlight the advantages of sequence-level importance sampling when properly designed.
All methods share the same data, sampling configuration, batch size, and number of training steps; FSPO differs only in employing log-space clipping with a length-scaled band.
For the \textbf{RLOO} baseline, we adopt the policy-loss formulation described in Appendix~\ref{app:prelim}, but we use the GRPO-style advantage $\hat A^{\text{GRPO}}$ rather than $\hat A^{\text{LOO}}$ for a fair comparison with the other three methods.

%% file: sections/results.tex
\section{Results and Analysis}
\label{sec:results}

\subsection{Main Results}
\label{sec:main-results}

\Cref{tab:main_results} reports results on \textsc{MATH500} (Avg@8) and \textsc{AIME24/25} (Avg@32) for two base model sizes. For each method we show the \emph{best} checkpoint (peak score across saved checkpoints) and the \emph{last} checkpoint (checkpoint of the last step). Overall, \textbf{FSPO} delivers consistent gains, with the largest margins on the harder AIME benchmarks and larger model size.

On \textbf{Qwen3-1.7B-Base}, \textbf{FSPO} attains the best AIME24 score (10.83/10.83) and the best last-average overall (29.16). On \textbf{Qwen3-8B-Base}, \textbf{FSPO} consistently outperforms the other methods across all benchmarks, achieving best/last averages of \textbf{49.79}/\textbf{48.98}, surpassing GRPO (+2.13/+1.93), RLOO (+1.99/+2.84), and GSPO (+2.82/+2.15). Gains are most pronounced on the more challenging AIME24 and AIME25: On \textsc{AIME24}, FSPO reaches \textbf{34.48/34.06}, yielding sizable gains versus GRPO (+3.23/+3.02), RLOO (+2.29/+4.06), and GSPO (+4.27/+3.85). On \textsc{AIME25}, FSPO achieves \textbf{24.69/24.69}, outperforming GRPO (+1.77/+2.19), RLOO (+1.67/+3.86), and GSPO (+2.19/+2.61).

Overall, gains of FSPO grow with model scale and task difficulty. This is expected as larger models and harder tasks induce broader, more heterogeneous response-length distributions, a regime where \textbf{FSPO}'s length-fair clipping yields the largest benefits.

\begin{table}[htbp]
\centering
\label{tab:main_results}
\setlength{\tabcolsep}{7pt}
\small
\begin{tabular}{lcccc}
\toprule
\textbf{Method} & \textbf{MATH500} & \textbf{AIME24} & \textbf{AIME25} & \textbf{Average} \\
& \scriptsize(Best/Last) & \scriptsize(Best/Last) & \scriptsize(Best/Last) & \scriptsize(Best/Last) \\
\midrule
\multicolumn{5}{l}{\textbf{Qwen3-1.7B-Base}}\\
base & 52.20& 3.02& 3.33& 19.52\\
GRPO  & 66.80/66.20  & 9.17/7.71 & 5.21/5.21 & 27.06/26.37\\
RLOO  & \textbf{70.80}/\textbf{70.80}  & 10.73/7.60 & \textbf{6.77}/\textbf{6.77} & \textbf{29.43}/28.39\\
GSPO  & 69.00/69.00  & 9.48/9.48 & 6.04/6.04 & 28.17/28.17\\
FSPO (ours) & 70.20/70.20 & \textbf{10.83}/\textbf{10.83} & 6.46/6.46 & 29.16/\textbf{29.16}\\
\midrule
\multicolumn{5}{l}{\textbf{Qwen3-8B-Base}}\\
base & 71.20& 10.00& 10.00& 30.40\\
GRPO  & 88.80/87.60& 31.25/31.04 & 22.92/22.50 & 47.66/47.05 \\
RLOO  & 88.20/87.60  & 32.19/30.00& 23.02/20.83& 47.80/46.14\\
GSPO  & 88.20/\textbf{88.20}& 30.21/30.21 & 22.50/22.08 & 46.97/46.83 \\
FSPO (ours) & \textbf{90.20}/\textbf{88.20}& \textbf{34.48}/\textbf{34.06} & \textbf{24.69}/\textbf{24.69} & \textbf{49.79}/\textbf{48.98} \\
\bottomrule
\end{tabular}
\caption{Performance across benchmarks. "base" indicates the performance of the starting base model without RL training. \textsc{MATH500} uses Avg@8; \textsc{AIME24}/\textsc{AIME25} use Avg@32. Each cell shows \textbf{Best/Last} results. Bold indicates the best within each column.}
\end{table}
\vspace{-4pt}
\subsection{Length-Fairness Diagnostics}
\label{sec:fair_diag}

\begin{figure}[htbp]
    \centering
    \includegraphics[width=0.75\linewidth]{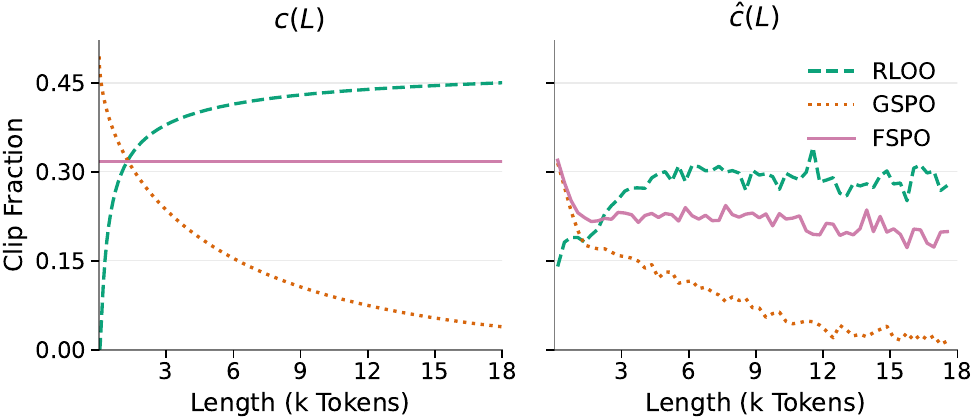}
    \caption{\textbf{Theoretical and empirical clip fraction.} \textbf{Left:} Theoretical clip probability \(c(L)\) computed from \Cref{eq:c_rloo,eq:c_gspo} using the hyperparameters in Appendix \ref{app:hyperparam_tuning}, where we set $\xi=\log(1+c_{\text{upper}})$.  \textbf{Right:} Observed clip fraction \(\hat c(L)\) with bin size \(=200\), collected from the experiments on Qwen3-8B-Base model.}
    \label{fig:qL}
\end{figure}

We examine the clip fraction as a function of response length and compare the
theoretical curve \(c(L)\) predicted by \eqref{eq:c_rloo} and \ref{eq:c_gspo};
see \Cref{fig:qL}. The observed clip fractions match the theoretical patterns, where \textbf{RLOO} clips more frequently as length increases especially on short to medium lengths, \textbf{GSPO} shows a clear decreasing trend with length, and \textbf{FSPO} remains comparatively flat across lengths. Slightly higher values in the shortest-length bins in FSPO are due
to limited samples and occasional outliers of abnormally short sequences.  

The scale gap between the theoretical and empirical curves is expected due to the asymmetry between the upper and lower clip ranges in implementation and the skew of positive vs. negative advantages: PPO’s pessimistic min surrogate effectively upper‑bounds clipping for positive‑advantage samples only (and vice versa).

For LRE, we compute the \emph{acceptance} rate \(q(L)=1-c(L)\) and exclude anomalously short cases with \(L<1000\).
The resulting LREs are \(0.162\) for RLOO, \(0.264\) for GSPO, and \(0.037\) for FSPO, where FSPO achieves the smallest LRE, according with its best performance demonstrated in 8B experiments.
\vspace{-2pt}
\subsection{Effectiveness: Learning Dynamics and Length Stability}
\begin{figure}[htbp]
    \centering
    \includegraphics[width=0.75\linewidth]{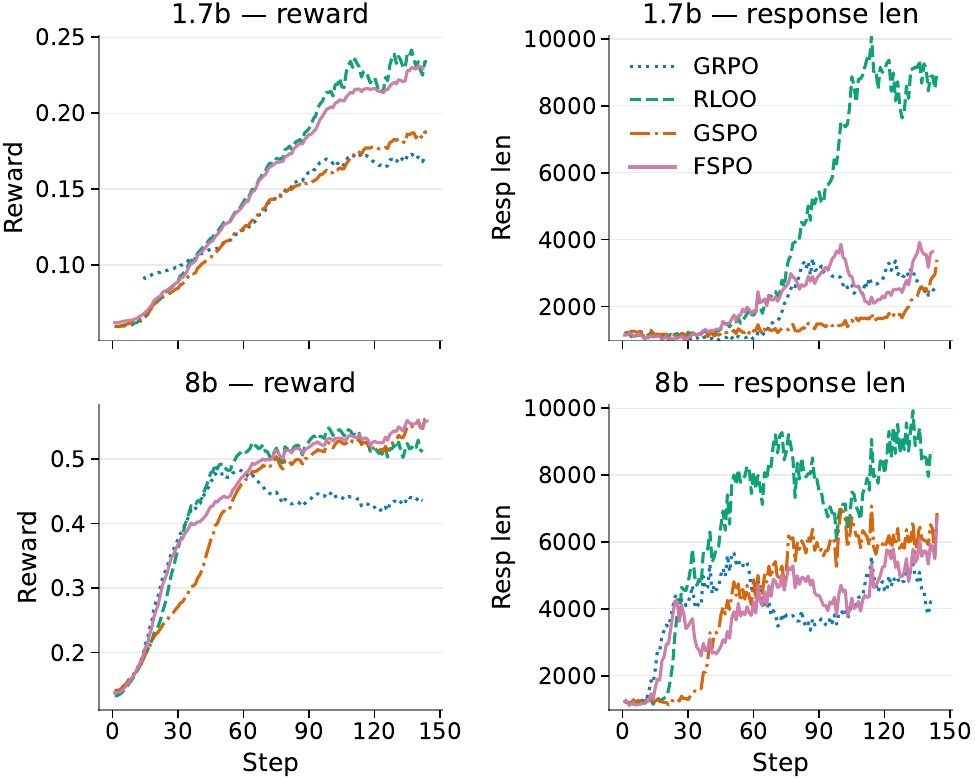}
    \caption{\textbf{Learning dynamics during training.} Left column: mean reward (1.7B and 8B). Right column: mean response length (1.7B and 8B). Reward curves are smoothed with EMA for visualization.}
    \label{fig:train_curve}
\end{figure}

As shown in \Cref{fig:train_curve}, both RLOO and FSPO learn quickly and increase response length early in training. However, RLOO’s response length later explodes to very large values, with much of the additional content being filler. A plausible explanation is that longer sequences are more likely to be clipped under RLOO; consequently, negative signals from long incorrect answers are suppressed, and the model fails to regulate length. Moreover, as responses grow longer, RLOO’s higher clip probability hampers learning and reward improvements plateau, whereas FSPO continues to make steady gains. By contrast, GSPO learns more slowly at the beginning and struggles to increase length, especially for the 1.7B model. On the 8B model, GSPO attains high rewards near or comparable to FSPO during training, yet its evaluation performance is suboptimal, indicating that length imbalance during training can impair calibration during generalization evaluation. FSPO attains the best performance with moderate average length on the 8B model, suggesting more balanced learning across lengths and more effective use of length. 
\vspace{-6pt}

To further assess downstream behavior, we report the \emph{overlong rate} (the proportion of samples that reach the maximum response length and are truncated) and mean response length after excluding overlong samples.
FSPO exhibits a markedly lower overlong rate, indicating stable control of response length. In contrast, methods with incorrect importance weights (GRPO, GSPO) show substantially higher overlong rates, while their mean lengths after excluding overlong samples are slightly lower, suggesting poor length allocation: responses tend to be either too long or too short, leading to suboptimal behavior. RLOO displays both higher overlong rate and larger length.
\begin{figure}[htbp]
    \centering
    \includegraphics[width=0.75\linewidth]{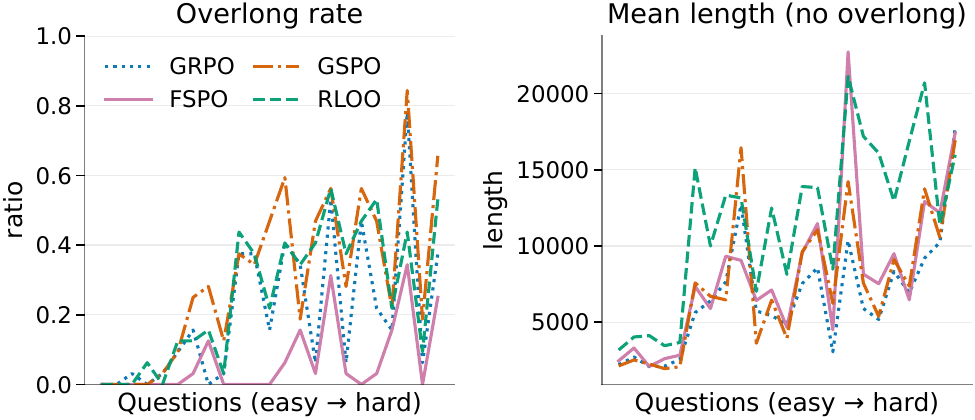}
    \caption{\textbf{Overlong rate and mean response length} on \textsc{AIME24}. We plot evaluation-time sampling; the x-axis orders the 30 problems from easy to hard, where difficulty is measured by the overall average accuracy across the four methods.}
    \label{fig:length_stability}
\end{figure}

\subsection{Ablation study: Larger Clip Range}
\begin{figure}[htbp]
    \centering
    \includegraphics[width=0.75\linewidth]{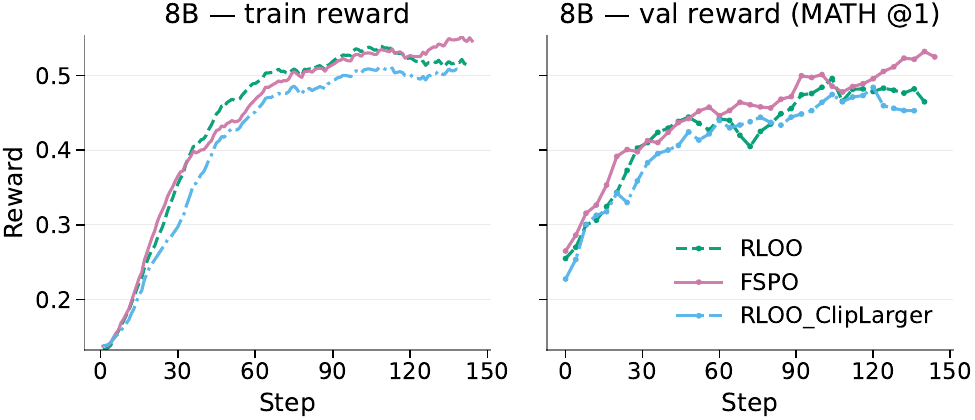}
    \caption{\textbf{Ablation: fixed larger clip range.} \textbf{Left:} mean rewards during training. \textbf{Right:} validation curves during training.}
    \label{fig:ablation}
\end{figure}

As in \Cref{fig:qL}, RLOO’s clip fraction is large, potentially due to its relatively small clip range (Appendix \ref{app:hyperparam_tuning}).
Note that in FSPO the \emph{ratio-level} clip range for a sequence with \(L=10{,}000\) is \(\exp(\sqrt{10000}\times 0.03)=20.09\), much larger than the \(1.667\) ($1 +c_{\text{upper}}$) used in RLOO. Thus, one may hypothesize that FSPO’s gains stem from being more permissive on long sequences than RLOO. To disentangle this, we evaluate RLOO with a \emph{fixed} larger clip range (upper \(=20\), lower \(=0.95\)). As shown in \Cref{fig:ablation}, this variant does not improve performance and can even be worse than standard RLOO. This indicates that \emph{length fairness}, rather than mere leniency toward long responses, is key to FSPO’s effectiveness.

%% file: sections/conclusion.tex
\section{Conclusion}
\label{sec:conclusion}
We studied the clipping mechanism in sequence-level importance sampling (IS) for RLVR scenarios, showing that a fixed clip range induces a length–reweighting pathology that biases acceptance across response lengths and distorts the effective objective. We formalized \emph{length fairness} via the Length Reweighting Error (LRE) and established a cosine–direction guarantee linking small LRE to update–direction fidelity. Guided by an approximate Gaussian law for the sequence log–IS sum, we proposed \textbf{FSPO}: clipping in log–IS space with a \(\sqrt{L}\)-scaled band that preserves IS semantics while equalizing acceptance across lengths. Empirically, on three math benchmarks and two model scales, FSPO flattens acceptance–by–length and delivers consistent gains, with the largest improvements on the 8B model. Looking ahead, we will extend evaluation to other RLVR settings (e.g. code, tool-use, etc.), develop and evaluate adaptive clip–range schedules, and combine FSPO with stronger advantage estimation to build more capable RL pipelines.

%% file: sections/appendix.tex
\appendix

\section{Preliminaries and Related Work}
\label{app:prelim}

\paragraph{Setup.}
Let $x_i\in\cX$ be a context (prompt) drawn from a data distribution $p(x)$,
and let $y_i = (y_{i,1}, ... y_{i, |y_i|})\in\mathcal{Y}$ be a response (a token sequence).
Under a policy $\pi_\theta$ (which is an LLM in our setting), the outcome distribution is
\[
\pi_{\theta}(y_i\mid x_i)\;=\;\prod_{t=1}^{|y_i|} \pi_\theta\big(y_{i,t} \mid h_{i,t}\big),
\]
where $h_{i,t}= (x_i,y_{i, <t})$ is the previous state.

\paragraph{RLHF and PPO.}
Reinforcement Learning from Human Feedback (RLHF) \cite{rlhf} frames alignment as policy optimization against a reward model learned from human preference data. RLHF utilizes PPO \cite{schulman2017ppo} algorithm: it assigns reward signal to the last token and uses GAE \cite{gae} to compute per-token advantages $\hat{A}_{i,t}$. The typical PPO objective is

\begin{equation}
\label{eq:ppo}
\mathcal{J}_{\mathrm{PPO}}(\theta)=
\frac{1}{N}\sum_{i = 1}^N\;\frac{1}{|y_i|}\sum_{t=1}^{|y_i|}
\min\!\Big(
r_{i,t}(\theta)\,\hat{A}_{i,t},\;
\mathrm{clip}\big(r_{i,t}(\theta),\,1-\epsilon,\,1+\epsilon\big)\,\hat{A}_{i,t}
\Big),
\end{equation}
\noindent where 
$r_{i,t}(\theta)=\dfrac{\pi_\theta(y_{i,t}\mid h_{i,t})}{\pi_{\theta_{\text{old}}}(y_{i,t}\mid h_{i,t})}$ is the token-level IS weight.

\paragraph{RLVR paradigm and GRPO.}
For math, programming and other verifiable tasks, recent systems adopt the RLVR \cite{lambert2025tulu3}  
paradigm: rule-based, sequence-level rewards that can be automatically checked. Representative methods include GRPO \cite{grpo}, DAPO \cite{dapo}, DrGRPO \cite{drgrpo}, etc. A typical GRPO-style update draws a group of $G$ completions $\{y_i\}_{i=1}^{G}$ for the same prompt $x$ under $\pi_{\theta_{\mathrm{old}}}$ (note here we abuse the notation of index $i$ as within-group index, while previous $i$ is the index in the dataset), computes sequence rewards $R_i=G(x,y_i)$, and uses the group mean as a baseline so that $\hat A_i = (R_i - \frac{1}{G}\sum_{j=1}^G R_j)/\sigma$, where $\sigma$ is the standard deviation of rewards of this group. The advantage for each token is $\hat A_{i,t} = \hat A_i$ for all $t = 1,2,...,|y_i|$. The objective of GRPO is similar to equation \ref{eq:ppo}, thus inherits the token-level IS weight and clipping.

\paragraph{Sequence-level Importance Sampling}
Recent works argue that LLM RL with sequence rewards should be performed with sequence-level importance sampling (IS). RLOO \cite{rloo} models LLM as a one-step bandit and treats a whole response as an action. According to the implementation of TRL \texttt{RLOO\_Trainer} \cite{trl_rloo_trainer}, the objective of RLOO is \begin{equation}
\label{eq:rloo}
\mathcal{J}_{\mathrm{RLOO}}(\theta)=
\mathbb{E}_{x,\{y_i\}\sim \pi_{\theta_{\mathrm{old}}}}\!\left[
\frac{1}{G}\sum_{i=1}^{G}
\min\!\Big(s_i(\theta)\,\hat A^{\mathrm{LOO}}_i,\;\mathrm{clip}\big(s_i(\theta),\,1-\epsilon,\,1+\epsilon\big)\,\hat A^{\mathrm{LOO}}_i\Big)
\right],
\end{equation}
where 
\begin{equation}
    \label{eq:seq_ratio}
    s_i(\theta)=\frac{\pi_{\theta}(y_i|x_i)}{\pi_{\theta_{\text{old}}}(y_i|x_i)}
\end{equation}
 is the sequence-level IS weight which matches the reward granularity, and $\hat A_i^{\text{LOO}}$ is the leave-one-out unbiased advantage estimator in \cite{rloo19}. A recent work GSPO \cite{gspo} pursue the same goal of sequence-level IS but normalize the IS ratio by length (e.g., using $s_i^{\mathrm{norm}}=\exp(\frac{1}{|y_i|}\log s_i)$) before clipping:
\begin{equation}
\label{eq:gspo}
\mathcal{J}_{\mathrm{GSPO}}(\theta)=
\mathbb{E}\!\left[
\frac{1}{G}\sum_{i=1}^{G}
\min\!\Big(s^{\mathrm{norm}}_i(\theta)\,\hat A_i,\;\mathrm{clip}\big(s^{\mathrm{norm}}_i(\theta),\,1-\epsilon,\,1+\epsilon\big)\,\hat A_i\Big)
\right].
\end{equation}
The idea of normalization by length is out of a similar motivation of providing a same scale for different lengths \cite{gspo}. However, we argue that this normalization does not reach the goal of balancing the clipping scale, and also undermines the correctness IS weights.

\section{Why Sequence-level IS for RLVR}
\label{app:why_seq}
RLOO \cite{rloo} models the entire generation as a single action (a bandit setting), but its context is RLHF and it does not yet discuss the inadequacy of token-level IS and the correctness of sequence-level IS for RLVR. GSPO \cite{gspo} notes that, in RLVR, the granularity of importance sampling should match the granularity of the reward, but it does not provide a detailed theoretical justification. 

Here we offer a detailed and illuminating discussion. 
As proved by \cite{approximate, schulman2015trpo} and presented in \cite{cs285}, the improvement of the objective between old and new parameters can be written as
\begin{align}
    J(\theta) - J(\theta_{old}) &= \EE_{\tau\sim \pi_{\theta}}\!\left[ \sum_{t=0}^{\infty}\gamma^t A^{\pi_{\theta_{old}}}(s_t, a_t)\right] \label{eq:J_diff_tau} \\
    &= \sum_{t=0}^{\infty}\EE_{s_t\sim p_{\theta}(s_t)}\!\left[\,\EE_{a_t\sim\pi_{\theta}(a_t|s_t)}\!\big[\gamma^t A^{\pi_{\theta_{old}}}(s_t,a_t)\big]\right] \label{eq:J_diff_sa} \\
    &= \sum_{t=0}^{\infty}\EE_{s_t\sim p_{\theta}(s_t)}\!\left[\,\EE_{a_t\sim\pi_{\theta_{old}}(a_t|s_t)}\!\bigg[\frac{\pi_{\theta}(a_t|s_t)}{\pi_{\theta_{old}}(a_t|s_t)}\,\gamma^t A^{\pi_{\theta_{old}}}(s_t,a_t)\bigg]\right].
\end{align}
This is where token-level importance sampling is introduced, as we need to express expectations under $\pi_\theta$ using samples sampled from $\theta_{old}$ in practice. Note that the state distribution $s_t\!\sim p_{\theta}(s_t)$ is not corrected by IS: $p(s_t)$ factors into a product over all previous actions, so naively correcting it can lead to high variance and poor estimates. In fact, the trust region in TRPO and the clipping mechanism in PPO are designed precisely to mitigate the mismatch incurred when estimating $s_t\!\sim p_{\theta}(s_t)$ using samples from $s_t\!\sim p_{\theta_{\mathrm{old}}}(s_t)$, since one can show that $p_{\theta_{\mathrm{old}}}(s_t)$ and $p_{\theta}(s_t)$ remain close when $\pi_{\theta_{\mathrm{old}}}$ and $\pi_{\theta}$ are close.

However, this formulation becomes problematic in the RLVR setting, where all tokens in a sequence share a single advantage. From \cref{eq:J_diff_tau} to \cref{eq:J_diff_sa}, the expectations over $(s_{t+1}, a_{t+1}, s_{t+2},\ldots)$ are marginalized out. That step requires the summand depends only on $(s_t,a_t)$ and not on the \emph{future} part of the trajectory. This condition fails in RLVR, in which $A(s_t,a_t)=A(\tau)$ for all $t$ in a sample sequence. 

To make this concrete, consider an imagined batch with two samples $y_a,y_b\sim \pi_{\theta_{\text{old}}}(y\mid x)$,
\[
y_a = (y_0, y_1, \ldots, y_t, y_{t+1}^{(a)}, y_{t+2}^{(a)}, \ldots),\qquad
y_b = (y_0, y_1, \ldots, y_t, y_{t+1}^{(b)}, y_{t+2}^{(b)}, \ldots),
\]
which share the same tokens up to index $t$ and diverge from $t{+}1$ onward. Suppose $y_a$ is correct and $y_b$ is incorrect (e.g., $A(y_a)=0.5$, $A(y_b)=-0.5$), $\pi_{\theta_{\text{old}}}(y_a)=\pi_{\theta_{\text{old}}}(y_b)$, and $\pi_{\theta}(y_a)>\pi_{\theta}(y_b)$. This setting implies that, under the current policy $\pi_{\theta}$, conditioned on the shared prefix $(y_0,\ldots,y_t)$, $y_a$ is more likely to occur than $y_b$, i.e., the model is more likely to answer correctly than incorrectly given the prefix. Intuitively, the next update should increase the likelihood of the prefix $(y_0,\ldots,y_t)$. Sequence-level IS achieves this, because
\[
\frac{\pi_{\theta}(y_a)}{\pi_{\theta_{\text{old}}}(y_a)} \;>\; \frac{\pi_{\theta}(y_b)}{\pi_{\theta_{\text{old}}}(y_b)},
\]
and the shared-prefix gradients are accordingly weighted more for $y_a$ than for $y_b$. By contrast, token-level IS assigns the same token-level ratios to the shared tokens $(y_0,\ldots,y_t)$ for both $y_a$ and $y_b$ (and for any other sample with that prefix), so it cannot express this desirable preference. This illustrates the advantage of sequence-level IS for RLVR.

Interestingly, in the case of sequence-level IS weight, the problem of $p_{\theta_{\text{old}}} \ne p_{\theta}$ as in PPO, TRPO does not hold, because the probability of the entire sequence is corrected. However, the clipping mechanism can be kept to solve the high-variance issue of sequence-level ratio and thus is still necessary.

\section{Proof of \Cref{thm:angle}}
\label{app:angle-proof}

\paragraph{Cosine lemma.}
For nonzero $u,v$, $\cos\angle(u,v)\ge\frac{\|v\|-\|u-v\|}{\|v\|+\|u-v\|}$.

\paragraph{Proof}
Note that $\cos\angle(g^b, g^*) = \cos\angle(g^b,\bar qg^*)$. To use cosine lemma, we want to bound $||g^b - \bar qg^*||$. 

Recall $g^b=\EE_L[q(L)g_L^{b}]$, $g^\star=\EE_L[g_L^\star]$, $\bar q=\EE[q(L)]$.
Decompose
\[
g^b-\bar q\,g^\star
=\EE_L\big[(q(L)-\bar q)\,g_L^\star\big]\;+\;\EE_L\big[q(L)\,(g_L^{b}-g_L^\star)\big].
\]
We bound its norm as
\begin{align*}
    \|g^b-\bar q g^\star\|
&\le \underbrace{\EE_L\!\big[\,|q(L)-\bar q|\,\|g^\star_L\|\,\big]}_{
\substack{\text{cross-length reweighting}}}
\;+\;
\underbrace{\EE_L\!\big[\,\|\,q(L)\,(g^{(b)}_L-g^\star_L)\|\,\big]}_{
\substack{\text{within-length stratification}}} \tag*{\textup{(by triangle inequality)}}\\
& \le \bar q\;\EE_L\left[\left|\frac{q(L)}{\bar q} - 1\right|||g^*_L||\right] + \eta\; \EE_L\left[||(q(L) - \bar q)g^*_L +\bar q\;g^*_L||\right] \tag*{\textup{(by \cref{ass:weak-strat})}}\\
& \le (1+\eta)\bar q\;\EE_L\left[\left|\frac{q(L)}{\bar q} - 1\right|||g^*_L||\right] + \eta\bar q\; \EE_L[||g^*_L||] \tag*{(by algebra)} \\
& \le \bar q\;(2\gamma(1+\eta)\; \mathrm{LRE } +\eta)\;\EE_L[||g^*_L||] \tag*{\textup{(by \cref{ass:cov} and definition of LRE)}}
\end{align*}

Recall that $\EE_L[||g^*_L||] = \kappa||g^*||$  and apply the cosine lemma with $u=g^b$, $v=\bar q g^*$, and \cref{thm:angle} is proved.

\paragraph{Weighted‑LRE variant}
If one wants to avoid any bounded co-variation concerns, define
\[
\mathrm{LRE}_w=\frac{1}{2}\EE\!\left[\left|\frac{q(L)}{\bar q}-1\right|\frac{\|g_L^\star\|}{\EE\|g_L^\star\|}\right].
\]
Then the same argument yields the bound with $\mathrm{LRE}$ replaced by $\mathrm{LRE}_w$.

\paragraph{More discussion: beyond length.}
It is worth noting that the proof of \Cref{thm:angle} does not fundamentally rely on the specific choice of $L$ as the partitioning variable. 
The argument only requires that the sample space can be partitioned into groups where (i) the signals across groups exhibit dispersion, and (ii) within-group stratification errors can be controlled. 
Thus, $L$ can be replaced by any other reasonable attribute that induces such a partition, for example bins of lengths (which justifies the binning process in our diagnostic plots), or other salient structural features of the data. 
This perspective suggests a more general criterion for designing clipping mechanisms: it should not introduce systematic bias across groups of any reasonable partitioning attribute, unless such a bias is intentionally desired.

\section{Configurations}
\label{app:config}

\subsection{Training Configurations}
\label{app:train_config}
We conduct experiments on a single 8 $\times$ \textsc{H}200 (140\,GB) GPUs. Under the configuration below, training for 1 epoch takes approximately 3--4 days; the wall-clock time grows with the average response length.

\begin{table}[h!]
\centering
\small
\caption{Training configuration.}
\label{tab:train_config}
\begin{tabular}{@{}ll@{}}
\toprule
\textbf{Item} & \textbf{Value} \\
\midrule
Prompt / Response max & 2{,}000 / 18{,}000 tokens \\
Global batch size (sequences) & 128 \\
mini-batch & 32 \\
per-GPU micro-batch & 32 \\
total steps & 144 \\
Optimizer \& LR & AdamW, $1{\times}10^{-6}$ \\
Parallelism & Megatron TP{=}8 \\
Rollout $n$ & 16 \\
vLLM GPU-util & 0.5 \\
Seeds & 42 \\
\bottomrule
\end{tabular}
\end{table}

\subsection{Algorithmic Hyperparameters and Tuning}
\label{app:hyperparam_tuning}

\textbf{Limitation}
Due to compute constraints, we did not perform an exhaustive hyperparameter search. For settings similar to ours, we recommend fixing the base clipping scale \(c\) at \(0.03\) or higher. However, this value may not transfer across substantially different datasets or model sizes.

\textbf{Interpreting \(c\), \(z\), and \(\hat\sigma\)}
In our formulation, the sequence log-IS ratio \(S\) is clipped by a band whose width is
\[
\text{band} \;=\; z \cdot \hat\sigma \,
\]
 Setting \(c=0.03\) is equivalent to choosing \(z{=}1\) and \(\hat\sigma{=}0.03\) in our environment (cf.\ Sec.~\ref{sec:gaussian}). In our experiments we first obtain \(\hat\sigma\) from a short baseline run, which inevitably introduces one preliminary pass.

To avoid a dedicated pilot run while preserving stability, we recommend:
\begin{enumerate}
    \item Initialize with \(z \in [1,\,1.5]\) and a heuristic \(\hat\sigma_0{=}0.03\) for warmup.
    \item Maintain a running estimate \(\hat\sigma_t\) of the log-IS ratio std via an exponential moving average (EMA): \(\hat\sigma_t \!\leftarrow\! (1-\alpha)\hat\sigma_{t-1} + \alpha\,\mathrm{std}_\text{batch}(S)\).
    \item Update the clip band as \(z \cdot \hat\sigma_t\).
\end{enumerate}

\begin{table}[h!]
\centering
\small
\caption{Algorithmic hyperparameters }
\label{tab:algo_hparams}
\begin{tabular}{@{}ll@{}}
\toprule
\textbf{Hyperparameter} & \textbf{Value}\\
\midrule
Upper Clip \(c_{\text{upper}}\) & 0.03  \\
Lower Clip \(c_{\text{lower}}\) & 0.03 \\
Dual Clip \(c_{\text{dual}}\) & 0.03 \\
Loss aggregation & token-mean  \\
use\_KL & disabled \\
Entropy coefficient & 0 \\
Advantage Estimator & GRPO-style \\
\bottomrule
\end{tabular}
\end{table}

\paragraph{Baseline Hyperparameters}

We also report the hyperparameters in our baseline implementations. Note that the clip-range here for baseline methods are in the ratio space as convention rather than the log space in our FSPO method. We adopt clip-higher as in \cite{dapo} for GRPO, and follow the guidance in GSPO paper for GSPO. The cliprange\_c hyperparameter is the parameter that VeRL uses to control dual-clip range.
\begin{table}[h!]
\centering
\small
\caption{Baseline clipping configuration}
\label{tab:baseline_clip}
\begin{tabular}{@{}llll@{}}
\toprule
\textbf{Baseline} & \textbf{cliprange\_upper} & \textbf{cliprange\_lower} & \textbf{cliprange\_dual} \\
\midrule
GRPO & 0.28 & 0.2 & 3.0 \\
RLOO & 0.667 & 0.4 & 3.0 \\
GSPO & 4e-4 & 3e-4 & disabled \\
\bottomrule
\end{tabular}
\end{table}

\subsection{Test-time Configurations}
\label{app:test-config}
We use OpenCompass \cite{opencompass} as our evaluation framework. 

\begin{table}[h!]
\centering
\small
\caption{Decoding configuration}
\label{tab:test_config}

\begin{tabular}{@{}ll@{}}
\toprule
\textbf{Item} & \textbf{Value} \\
\midrule
Temperature & 0.6 \\
Top-$p$  & 0.95 \\
Top-$k$ & 200 \\
Max generation tokens & 32{,}000 \\
Batch size & 256 \\
Tensor parallel & 8 \\
Data parallel & 1 \\
\bottomrule
\end{tabular}
\end{table}